\pdfoutput=1

\documentclass[11pt]{article}

\usepackage[]{ACL2023}

\usepackage{times}
\usepackage{latexsym}

\usepackage[T1]{fontenc}

\usepackage[utf8]{inputenc}

\usepackage{microtype}

\usepackage{inconsolata}

\usepackage{graphicx}
\usepackage{amsmath}
\usepackage{amssymb}
\usepackage{booktabs}
\usepackage{bbm}
\usepackage{enumitem}
\usepackage{multirow}
\usepackage{subcaption}

\title{Sample Attackability in Natural Language Adversarial Attacks}

\author{Vyas Raina \\
  University of Cambridge \\
  \texttt{vr313@cam.ac.uk} \\\And
  Mark Gales \\
  University of Cambridge \\
  \texttt{mjfg@cam.ac.uk} \\}

\begin{document}
\maketitle
\begin{abstract}
Adversarial attack research in natural language processing (NLP) has made significant progress in designing powerful attack methods and defence approaches. However, few efforts have sought to identify which source samples are the most attackable or robust, i.e. can we determine for an unseen target model, which samples are the most vulnerable to an adversarial attack. This work formally extends the definition of sample attackability/robustness for NLP attacks. Experiments on two popular NLP datasets, four state of the art models and four different NLP adversarial attack methods, demonstrate that sample uncertainty is insufficient for describing characteristics of attackable/robust samples and hence a deep learning based detector can perform much better at identifying the most attackable and robust samples for an unseen target model. Nevertheless, further analysis finds that there is little agreement in which samples are considered the most attackable/robust across different NLP attack methods, explaining a lack of portability of attackability detection methods across attack methods.~\footnote{Code: \url{https://github.com/rainavyas/nlp_attackability}}
\end{abstract}

\section{Introduction}

With the emergence of the Transformer architecture~\citep{DBLP:journals/corr/VaswaniSPUJGKP17}, natural language processing (NLP) models have demonstrated impressive performance in many tasks, ranging from standard sentiment classification~\citep{10.1145/3548772} to summarisation~\citep{9183355} and translation~\citep{DBLP:journals/corr/abs-2002-07526}. However, \citet{https://doi.org/10.48550/arxiv.1412.6572} demonstrated that deep learning models are susceptible to adversarial attacks, where carefully crafted small imperceptible changes applied to original, natural inputs can cause models to mis-classify. In response, extensive efforts have explored methods to combat the threat of adversarial attacks by training with adversarial examples~\citep{qian2022survey} or building separate detection systems~\citep{DBLP:journals/corr/abs-2103-03000, Raina_2022}. However, little or no work has sought to determine which input samples are the most susceptible to adversarial attacks. Are certain input samples easier to adversarially attack and if so can we efficiently identify these \textit{attackable} samples? The ability to identify the \textit{attackable} and in converse the \textit{robust} samples has applications in a range of sample-selection tasks. For example, in the field of active learning~\citep{5581075}, the query system can be designed to select the most \textit{attackable} samples. Similarly, knowledge of sample attackability is useful for weighted adversarial training~\citep{Kim2021EntropyWA}, where the aim is to augment the training set with only the most \textit{useful} adversarial examples.

In the image domain, \citet{raina2023identifying} formally define the notion of sample attackability as the minimum perturbation size required to change a sample's output prediction from the target model. Running iterative adversarial attacks to determine this minimum perturbation size for a single sample is inefficient. \citet{Kim2021EntropyWA} use entropy (uncertainty) as a proxy function for sample attackability, but, \citet{raina2023identifying} demonstrate that training a deep learning based classifier to predict the most attackable samples (and most robust samples) is the most effective method in the image domain. Therefore, this works extends the use of a deep learning based system to identify the most attackable and robust samples in NLP tasks. As a measure of a sample's attackability, it is challenging to define a sample's \textit{perturbation size} for natural language. Following \citet{raina2023identifying} in the image domain, this work uses the \textit{imperceptibility} threshold in the definition of an adversarial attack as a measure of the perturbation size. To align with human perception, imperceptibility constraints for NLP aim to limit the semantic change in the text after an adversarial attack. These imperceptibility constraints can be grouped into two stages: 1) pre-transformation constraints (e.g. no stopword changes) that limit the set of acceptable adversarial examples; and 2) distance constraints that only allow for a subset of the acceptable adversarial examples, where the distance constraint explicitly restricts the \textit{distance moved} by an adversarial example from the original example to satisfy a specified imperceptibility threshold. This distance can be measured for example using the Universal Sentence Encoder~\citep{herel2022preserving}. A sample subject to a specific NLP attack method (with defined pre-transformation constraints) will have an associated set of acceptable adversarial examples. The attackability of the sample can thus be given by the smallest distance constraint imperceptibility threshold that at least one acceptable adversarial example in the set satisfies.

Default imperceptibility thresholds for the distance constraints proposed for NLP attack methods can often lead to unnatural adversarial examples~\citep{morris-etal-2020-reevaluating}. Hence, in this work, we use separate thresholds for defining \textit{attackable} and \textit{robust} samples. A sample's minimum perturbation size is required to be within a much stricter imperceptibility threshold to be termed \textit{attackable}, whilst in converse a sample's minimum perturbation size has to be greater than a more generous imperceptibility threshold to be termed \textit{robust}. The deep learning based attackability classifier proposed in \citet{raina2023identifying} is successfully used to identify the attackable and robust samples for unseen data and unseen target models. However, in contrast to the image domain, it is found in NLP that the trained attackability detector fails to determine the attackable samples for different unseen NLP attack methods. This work extensively analyzes this observation and offers an explanation rooted in the inconsistency of imperceptibility definitions for different NLP attack methods.

\section{Related Work}
In the image domain \citet{DBLP:journals/corr/abs-2010-12989} introduce the notion of sample attackability through the language of \textit{vulnerability} of a sample to an adversarial attack. This vulnerability is abstractly defined as the distance of a sample to a model's decision boundary. \citet{raina2023identifying} offer a more formal and extensive estimate of a sample's vulnerability/attackability by considering the smallest perturbation size, aligned with an adversarial attack's imperceptibility measure, to change a sample's class prediction. Other research in the field of \textit{weighted adversarial training}~\citep{Kim2021EntropyWA}, has also implicitly considered the notion of sample attackability. The aim in weighted adversarial training is train with the more \textit{useful} adversarial examples, which are arguably sourced from the more \textit{attackable} original samples. For example \citet{Kim2021EntropyWA} use model entropy to estimate this attackability, whilst \citet{DBLP:journals/corr/abs-2010-12989} use model confidence and \citet{raina2023identifying} are successful in using a deep-learning based estimator of attackability. In the field of NLP, little work has explored weighted adversarial training. \citet{xu-etal-2022-towards} propose a meta-learning algorithm to lean the importance of each adversarial example, but this has no direct relation to a source sample's attackability. Finally, in the field of active learning~\citep{DBLP:journals/corr/abs-2009-00236, 5581075} there has also been implicit consideration of adversarial perturbation sizes as a measure of a sample's value. The aim in active learning is to select the most useful subset of samples in a dataset to train a model on. In the image domain, \citet{DBLP:journals/corr/abs-1802-09841} propose the use of the smallest adversarial perturbation size for each sample to measure the distance to the decision boundary. However, there is no explicit consideration of sample attackability or design of an efficient method to identify the attackable samples.

\section{Adversarial Attacks} \label{sec:attack}

In both the image and NLP domain, an untargeted adversarial attack is able to fool a classification system, $\mathcal F()$, by perturbing an input sample, $\mathbf x$ to generate an adversarial example $\tilde{\mathbf x}$ to cause a change in the output class,
\begin{equation}
    \mathcal F(\mathbf x)\neq\mathcal F(\mathbf{\tilde x}).
\end{equation}
It is necessary for adversarial attacks to be \textit{imperceptible}, such that adversarial examples, $\mathbf{\tilde x}$ are not easily detectable/noticeable by humans. It is inefficient and expensive to rely on manual human measures of attack imperceptibility, so instead proxy measures are used to enforce imperceptibility of an adversarial attack. For images, the $l_p$ norm is considered a good proxy for human perception of imperceptibility. However, in NLP it is more challenging to ensure imperceptibility. Despite earlier research introducing only visual constraints on the adversarial attacks~\citep{goyal2023survey, DBLP:journals/corr/abs-1801-04354, DBLP:journals/corr/abs-1712-06751, DBLP:journals/corr/abs-1905-11268, tan-etal-2020-morphin, DBLP:journals/corr/abs-1812-05271}, e.g. number of words changed as per the Levenshtein distance, recent research considers more sophisticated measures seeking to measure the \textit{semantic} change in text sequences~\citep{DBLP:journals/corr/abs-2009-07502, DBLP:journals/corr/abs-1907-11932, ren-etal-2019-generating, DBLP:journals/corr/abs-1909-06723, garg-ramakrishnan-2020-bae, DBLP:journals/corr/abs-1804-07998, DBLP:journals/corr/abs-2004-09984}. In general, modern NLP imperceptibility constraints can be separated into two stages: \textit{pre-transformation} constraints and \textit{distance} constraints. Pre-transformation constraints typically limit the attack mechanism to encourage little change in semantic content. For example, stop-word transformations will be prevented or any word substitutions will be restricted to appropriate synonyms. A collection of pre-transformation constraints, as specified by a particular attack method, limit the available set, $\mathcal A$ of possible adversarial examples that can be considered for a specific sample, $\mathbf{x}$, such that
\begin{equation}\label{eqn:pre}
    \mathbf{\tilde x}\in\mathcal A.
\end{equation}
The \textit{distance}-based constraints are further constraints that explicitly aim to limit the \textit{distance} between the original sample $\mathbf x$ and the adversarial example, $\mathbf{\tilde x}$ to ensure a small perceived semantic change. This \textit{distance} can be measured via a proxy function, $\mathcal G$,
\begin{equation}\label{eqn:const}
    \mathcal{G}(\mathbf x, \mathbf{\tilde x}) \leq\epsilon, 
\end{equation}
where $\epsilon$ represents the maximum imperceptibility threshold. A popular example of such a distance constraint is a limit on the cosine-distance in a sentence embedding space, e.g.,
\begin{equation}
    \mathcal{G}(\mathbf x, \mathbf{\tilde x}) = 1-\mathbf{h}^T{\mathbf{\tilde h}},
\end{equation}
where $\mathbf h$ and $\mathbf{\tilde h}$ are the normalized vector embedding representations of the word sequences $\mathbf x$ and $\mathbf{\tilde x}$.



\section{Sample Attackability Definition}

Sample attackability is concerned with how \textit{easy} it is to adversarially attack a specific sample. The notion of sample attackability is formally introduced by \citet{raina2023identifying}, where a specific input sample, $\mathbf x_n$'s attackability for a specific model, $\mathcal F_k$ is given by the theoretical minimum perturbation size, $\hat{{\delta}}_n^{(k)}$ within which a sample can be successfully attacked. However, it is not simple to define the perturbation size for an adversarial attack in NLP. The simplest definition for the perturbation size, ${\delta}$, for a specific attack method with a specific set of acceptable adversarial examples, $\mathcal A$ (Equation \ref{eqn:pre}), is to use the distance-based proxy function, $\mathcal{G}$ (Equation \ref{eqn:const}), such that ${\delta}=\mathcal G(\mathbf x, \mathbf{\tilde x})$. Then the minimum perturbation size, $\hat{{\delta}}_n^{(k)}$ for sample $n$ and model $k$ is,
\begin{equation}\label{eqn:nlp-pert}
    \hat{{\delta}}_n^{(k)}  = \min_{\substack{\mathbf x\in\mathcal A,\\ \mathcal F_k(\mathbf x_n)\neq\mathcal  F_k(\mathbf x)}}\left\{\mathcal G(\mathbf x_n, \mathbf x)\right\}.
\end{equation}
We aim to use a sample's minimum perturbation size to classify it as \textit{attackable}, \textit{robust} or neither. Default distance-based imperceptibility constraints defined using $\mathcal G$ for various NLP attack methods can lead to unnatural adversarial examples and so we use separate and stricter thresholds for classifying samples as \textit{attackable} or \textit{robust}. Hence, as in \citet{raina2023identifying}, we define sample $n$ as \textit{attackable} for model $k$ if the smallest adversarial perturbation is less than a strict threshold,  $\mathbf{A}_{n,k} = ({\hat\delta}^{(k)}_n<\epsilon_a)$, where any sample that is not attackable can be denoted as $\Bar{\mathbf{A}}_{n,k}$. Conversely, a sample is defined as \textit{robust}, if its adversarial perturbation size is larger than a separate, but more generous (larger) set threshold, $\mathbf{R}_{n,k}=(\hat\delta^{(k)}_n>\epsilon_r)$.

It is informative to identify samples that are \textit{universally} attackable/robust across different models. We can thus extend the definition for \textit{universality} as follows. A sample, $n$, is \textbf{universally attackable} if,
\begin{equation}\label{eqn:uni-att}
    \mathbf A_{n}^{(\mathcal M)} = \bigcap_{k, \mathcal F_k\in\mathcal M} \hspace{0.5em} \mathbf{A}_{n,k},
\end{equation}
where $\mathcal M$ is the set of models in consideration. Similarly a sample is \textbf{universally robust} if, $\mathbf R_{n}^{(\mathcal M)}=\bigcap_{k, \mathcal F_k\in\mathcal M} \hspace{0.5em}\mathbf{R}_{n,k}$. Note that all of the attackability definitions in this section are for a specific attack method (e.g. Textfooler), as definition of the perturbation size in Equation \ref{eqn:nlp-pert} uses the distance-based imperceptibility constraint, $\mathcal G$ specific to an attack method. Portability of these definitions and attackability detection models across attack methods is explored in Section \ref{sec:exp-port}.

\section{Attackability Detector} \label{sec:design}

The definition of attackable and robust samples uses the minimum perturbation size (as per a distance-based constraint) for an NLP adversarial attack on a sample. When trying to determine which samples are \textit{attackable}, it is slow and expensive to run an adversarial attack iteratively to find the minimum perturbation size. Further, often one may not have access to an unseen target model, $\mathcal F_t$ or even the target sample, $n$ to perform an adversarial attack upon. Hence, in this setting, it is necessary to have a simple and efficient process that can determine whether samples in an unseen dataset are attackable for an unseen target model. Inspired by \citet{raina2023identifying}, this section describes a method to train a simple deep-learning attackability detector to identify the attackable and robust samples in an unseen dataset, for an unseen target model, $\mathcal F_t$. We give the deep-learning attackability detector access to a seen dataset, $\{\mathbf x_n, y_n\}_{n=1}^N$ and a set of seen models, $\mathcal M = \{\mathcal F_1, \hdots, \mathcal F_{|\mathcal M|}\}$, such that $\mathcal F_t\notin\mathcal M$. Each model can be represented as an encoder embedding stage, followed by a classification stage,
\begin{equation}
    \mathcal{F}_k(\mathbf x_n) = \mathcal F_k^{(\texttt{cl})}(\mathbf h_{n,k}),
\end{equation}
where $\mathbf h_{n,k}$ is the model encoder's embedding of $\mathbf x_n$. For each seen model in $\mathcal M$, a separate attackability detector can be trained. For a specific seen model, $k$, we can measure the attackability of each sample using the minimum perturbation size (Equation \ref{eqn:nlp-pert}), $\{{\hat\delta}^{(k)}_n\}_{n=1}^N$. It is most efficient to exploit the encoder embedding representation of input text sequences, $\mathbf h_{n,k}$, already learnt by each model. Hence, each deep attackability detector, $\mathcal D_{\theta}^{(k)}$, with parameters $\theta$, can be trained as a binary classification task to determine the probability of a sample being attackable for model $k$, using the encoder embedding at the input,
\begin{equation}
    p(\mathbf{A}_{n,k}) = \mathcal D_{\theta}^{(k)}(\mathbf h_{n,k}).
\end{equation}
 Consistent with \citet{raina2023identifying}, we use a simple, single hidden-layer fully connected network architecture for each attackability detector, $\mathcal D$, such that,
\begin{equation}\label{eqn:fcn}
    \mathcal D_{\theta}(\mathbf h) = \sigma(\mathbf{W}_1\sigma(\mathbf W_0\mathbf h)),
\end{equation}
where $\mathbf{W}_0$ and $\mathbf{W_1}$ are the trainable parameters and $\sigma()$ is a standard sigmoid function. This collection of model-specific detectors can be used to estimate the probability of a new sample being attackable for an unseen target model, $\mathcal F_t$. It is most intuitive to take an expectation over the seen model-specific detector attackability probabilities, 
\begin{equation}
    p(\mathbf{A}_{n,t})\approx \frac{1}{|\mathcal M|}\sum_{k, \mathcal F_k\in\mathcal M}p(\mathbf A_{n,k}).
\end{equation}
\citet{raina2023identifying} demonstrated that this estimate in the image domain does not capture the samples that are attackable specifically for the target model, $\mathcal F_t$'s specific realisation. Therefore, we seek instead to estimate the probability of a \textit{universally attackable} sample (defined in Equation \ref{eqn:uni-att}),
\begin{equation}\label{eqn:uni-det}
    p(\mathbf{A}^{(\mathcal M+t)}_{n})\approx \left[\frac{1}{|\mathcal M|}\sum_{k, \mathcal F_k\in\mathcal M}p(\mathbf A_{n,k})\right ]^{\alpha(\mathcal M)},
\end{equation}
where the parameter $\alpha(\mathcal M)$ models the idea that the probability of sample being universally attackable should decrease with the number of models (note that this is empirically observed in Figure \ref{fig:sweep}). An identical approach can be used to train detectors to give the probability of a sample being \textit{universally robust}, $p(\mathbf{R}^{(\mathcal M+t)}_{n})$. 

 The attackability/robustness of samples can also be estimated using simple uncertainty based approaches, such as entropy~\citep{Kim2021EntropyWA} or a sample's class margin measured by model confidence~\citep{DBLP:journals/corr/abs-2010-12989}. These uncertainty measures can then also be compared to strict thresholds to classify samples as attackable or robust. Experiments in Section \ref{sec:exp} compare the deep-learning based attackability detector to uncertainty-based attackability detectors. To assess which attackability detector performs the best in identifying attackable samples for the unseen target model, $\mathcal F_t\notin\mathcal M$, we consider four variations on defining a sample, $n$ as attackable~\citep{raina2023identifying}.\newline
 \noindent \textbf{all}- the sample is attackable for the unseen target model.
\begin{equation}\label{eqn:all}
      \mathbf{A}_{n,t} = ({\hat\delta}^{(t)}_n < \epsilon_a).
    \end{equation}
\noindent\textbf{uni} - the sample is universally attackable for the seen models and the unseen target model.
    \begin{equation}\label{eqn:uni}
        \mathbf{A}^{(\mathcal M+t)}_{n} = \mathbf{A}_{n,t} \cap \mathbf{A}^{(\mathcal M)}_{n}.
    \end{equation}
\noindent\textbf{spec} - the sample is attackable for the target model but not universally attackable for the seen models.
    \begin{equation}\label{eqn:spec}
       \mathbf A^{\texttt{spec}}_{n,t} = \mathbf A_{n,t}\cap\bar{\mathbf{A}}^{(\mathcal M)}_{n}.  
    \end{equation}
 \noindent\textbf{vspec} - a sample is specifically attackable for the unseen target model only.
    \begin{equation}\label{eqn:vspec}
        \mathbf A^{\texttt{vspec}}_{n,t} = \mathbf{A}_{n,t} \cap \left(\bigcap_{k, \mathcal F_k\in\mathcal M} \hspace{0.5em} \bar{\mathbf{A}}_{n,k}\right).
    \end{equation}

Given that the deep learning based attackability detectors are trained to identify universally attackable samples (Equation \ref{eqn:uni-det}), they are expected to perform best in the \textit{uni} evaluation setting.

The corpus-level performance of an attackability detector for an unseen dataset can be reported using precision and recall. A selected threshold, $\beta$, is used to class the output of detectors, e.g. $p(\mathbf{A}^{(\mathcal M+t)}_{n})>\beta$ classes sample $n$ as attackable. The precision is $\texttt{prec}=\text{TP}/\text{TP+FP}$ and recall is $\texttt{rec}=\text{TP}/\text{TP+FN}$, where FP, TP and FN are standard counts for False-Positive, True-Positive and False-Negative. An overall score is given with the F1-score, $\text{F1} = \text{2}*(\texttt{prec}*\texttt{rec})/(\texttt{prec}+\texttt{rec})$. By sweeping the threshold $\beta$ a full precision-recall curve can be generated and typically the threshold with the greatest F1-score is selected as an appropriate operating point.

\section{Experiments} \label{sec:exp}

\subsection{Setup} \label{sec:setup}

Experiments in this section aim to understand how well a deep-learning based detector, described in Section \ref{sec:design}, performs in identifying attackable samples for an unseen dataset and an unseen target model, $\mathcal F_t$, where the detector only has access to a separate set of \textit{seen} models, $\mathcal M$ during training. There are equivalent experiments looking to detect the most robust samples too. The performance of the deep learning based detector is compared to a baseline of uncertainty-based detectors (model confidence), inspired by \citet{DBLP:journals/corr/abs-2010-12989}, in which the samples with the most uncertain model predictions are identified as attackable and in converse the most certain samples are deemed to be robust. Specifically, two forms of uncertainty-based detectors are considered: 1) conf-u, where there is no access to the confidence from the unseen target model and so a sample's uncertainty is measured by an average of the confidence of the seen models, $\mathcal M$; and as a realistic reference we also consider 2) conf-s, where there is access to the target model output such that the target model's confidence is used directly as a measure of sample uncertainty.

Two popular natural language classification datasets are used in these experiments. First, the Stanford Sentiment Treebank2 dataset (sst)~\citep{socher-etal-2013-recursive} is a movie review dataset with each review labelled as positive or negative. There are 6920 training samples, 872 validation samples and 1820 test samples. We also consider the Twitter Emotions dataset~\citep{saravia-etal-2018-carer}, which categorizes tweets into one of six emotions: sadness, love, anger, surprise, joy and fear. This dataset contains 16,000 training samples, 2000 validation samples and 2000 test samples. For training of the attackability detectors, access was provided to only the validation data and hence the test data was used as an unseen set of samples to assess the performance of attackable sample detection.

These experiments work with four state of the art NLP transformer-based models: BERT (bert)~\citep{DBLP:journals/corr/abs-1810-04805}, XLNet (xlnet)~\citep{DBLP:journals/corr/abs-1906-08237}, RoBERTa (roberta)~\citep{DBLP:journals/corr/abs-1907-11692} and Electra (electra)~\citep{DBLP:journals/corr/abs-2003-10555}. Each model is of \textit{base}-size (110M parameters). Finetuning on sst and twitter used ADAMW optimizer, 3 epochs and a learning rate of 1e-5. The performance of the models is given in Table \ref{tab:acc}. Three models (bert, xlnet, roberta) are treated as \textit{seen} models, $\mathcal M$, that the attackability detector has access to during training. The electra model is maintained as the \textit{unseen} target model, $\mathcal F_t\notin \mathcal M$ used only to assess the performance of the attackability detector.

\begin{table}[htb!]
    \centering
    \small
    \begin{tabular}{lrr}
    \toprule
        Model & sst & twitter \\ \midrule
        bert&  91.8& 92.9\\
        xlnet & 93.6 & 92.3\\
        roberta & 94.7 &93.4 \\ \midrule
        electra & 94.7 &93.3\\
        \bottomrule
    \end{tabular}
    \caption{Model Accuracy (\%)}
    \label{tab:acc}
\end{table}

Four adversarial attack types are considered in these experiments: Textfooler (tf)~\citep{DBLP:journals/corr/abs-1907-11932}, Bert-based Adversarial Examples (bae)~\citep{garg-ramakrishnan-2020-bae}, Improved Genetic Algorithm (iga)~\citep{DBLP:journals/corr/abs-1909-06723} and Probability Weighted Word Saliency (pwws)~\citep{ren-etal-2019-generating}. In the bae attack we consider specifically the BAE-R attack mode from the paper, where the aim is to \textbf{r}eplace tokens. For NLP adversarial attacks Section \ref{sec:attack} discusses the nature of imperceptibility constraints, where constraints can either be \textit{pre-transformation} constraints (Equation \ref{eqn:pre}) or \textit{distance}-based constraints (Equation \ref{eqn:const}). Table \ref{tab:constraints} summarises the constraints for each of the selected attack methods in this work. In the attackability detection experiments, the textfooler attack is treated as a \textit{known} attack type, which the attackability detector has knowledge of during training, whilst the bae attack is an \textit{unknown} attack type, reserved for evaluation of the detector to assess the portability of the detector across attack methods. Evaluation of the attackability detector on the unseen datasets and the unseen target model (electra) with samples attacked by the known textfooler attack is referred to as \textit{matched} evaluation, whilst samples attacked by the unknown bae attack is referred to as \textit{unmatched} evaluation. The final two attack methods, pwws and iga, are used to further explore portability across attack methods in Section \ref{sec:exp-port}.

\begin{table}[htb!]
    \centering
    \small
    \begin{tabular}{l|cccc}
    \toprule
        constraints & tf & bae & pwws & iga\\ \midrule
       no repeat tkn changes& \checkmark&\checkmark&\checkmark\\
       no stopword changes  & \checkmark&\checkmark&\checkmark&\checkmark\\
       same part of speech swaps & \checkmark &\checkmark\\
       nearest neighbour syns swap & \checkmark & &&\checkmark\\
       language model syns swap & & \checkmark& \\
       wordnet syns swap & & & \checkmark\\
       
 \midrule\midrule
       Universal Sentence Encoding &\checkmark &\checkmark \\ 
       Word Embedding Distance  & \checkmark&&&\checkmark\\ 
       \% of words changed & & & & \checkmark\\
           \bottomrule   
    \end{tabular}
    \caption{Pre-transformation (top) and Distance-based (bottom) constraints for nlp adversarial attack methods.}
    \label{tab:constraints}
\end{table}

\subsection{Results}

\begin{figure}[htb!]
    \centering
    \includegraphics[width=\linewidth]{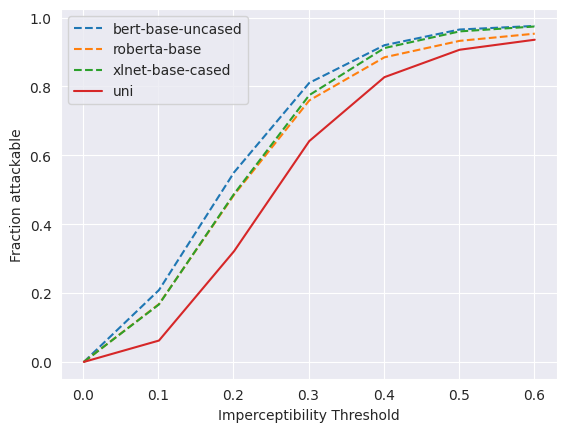}
    \caption{Fraction of \textit{attackable} samples.}
    \label{fig:sweep}
\end{figure}

The first set of experiments consider the \textit{matched} setting, where the known tf attack method is available at training time for the attackability detectors and also used to evaluate the attackability detectors. For each seen model, $\mathcal M$ (bert, xlnet, roberta), the tf attack method is used to determine the minimum perturbation size (as per distance-based constraints of the NLP attack method), $\hat\delta^{(k)}_n$, required to successfully attack each sample, $n$ in the validation dataset (Equation \ref{eqn:nlp-pert}). Note from Table \ref{tab:constraints} that this perturbation size is measured using the cosine distance for both word embeddings and Universal Sentence Encoder embeddings for the tf attack method. Using the sst data as an example, Figure \ref{fig:sweep} shows the fraction of samples, $f$ that are successfully attacked for each model, as the adversarial attack constraint, $\epsilon_a$ is swept: $f=\frac{1}{N}\sum_n^N\mathbbm{1}_{\mathbf A_{n,k}}$. Given this plot, we can sensibly define strict thresholds for attackability and robustness for the tf attack method: samples with a perturbation size below $\epsilon_a=0.15$ are termed \textit{attackable} and samples with a perturbation size above $\epsilon_r=0.35$ are termed \textit{robust}.

The aim now is to identify the attackable samples in the unseen test dataset that are vulnerable to attack as per the tf attack method for an unseen target model, $\mathcal F_t$ (electra). As described in Section \ref{sec:setup}, two baseline methods are considered: conf-u, which has no knowledge of the target electra model and so uses the average confidence from the seen models, $\mathcal M$ (bert, xlnet and roberta); and conf-s, which has access to the predictions from the target model and so explicitly uses the target model's confidence to identify attackable samples. The method of interest in this work is the \textit{deep}-learning based detector described in Section \ref{sec:design}. Here, a single layer fully connected network (Equation \ref{eqn:fcn}) is trained with \textit{seen} (bert, xlnet, roberta) model's final layer embeddings, using the validation samples in a binary classification setting to detect attackable samples. The number of hidden layer nodes for each model's FCN is set to the encoder embedding size of 768. Training of the FCNs used a batch-size of 8, 5 epochs, a learning rate of 1e-5 and an ADAMW optimizer. Table \ref{tab:m-a} shows the (best) F1 scores for detecting \textit{attackable} samples on the unseen test data for the unseen target electra model, in the matched setting. Note that the scale of F1 scores can vary significantly between evaluation settings (\textit{spec}, \textit{vspec}, \textit{uni} and \textit{all}) as the prevalence of samples defined as \textit{attackable} in a dataset are different for each setting and so it is not meaningful to compare across evaluation settings. Table \ref{tab:m-r} presents the equivalent results for detecting robust samples, where the definitions for each evaluation setting update to identifying \textit{robust} samples ($\mathbf R_{n,k}$). For both the twitter and sst datasets, in detecting attackable samples, the deep detection method performs best in all evaluation settings, whilst for robust sample detection it performs significantly better in only the \textit{uni} evaluation setting. Better performance in the \textit{uni} setting is expected due to the deep detection method having been designed explicitly to detect universally attackable samples (across models) (Equation \ref{eqn:uni-det}), whilst for example the \textit{conf-s} detection method has direct access to the target unseen model (electra) and so has the ability to perform competitively in the \textit{spec} and \textit{vspec} settings.

\begin{table}[htb!]
    \centering
    \small
    \begin{tabular}{ll|ccc}
    \toprule
       Setting  & &conf-s & conf-u & deep  \\ \midrule
        \multirow{2}{*}{all} & sst & 0.244&0.243&0.461\\
        & twitter & 0.457 &0.457&0.516\\ \midrule
        \multirow{2}{*}{uni} & sst &0.103&0.110&0.281\\\
        & twitter& 0.299&0.300 &0.435\\ \midrule
        \multirow{2}{*}{spec} & sst &0.165&0.165&0.130\\
        & twitter& 0.220& 0.222&0.236\\ \midrule
        \multirow{2}{*}{vspec} & sst &0.038&0.047&0.052\\
        & twitter& 0.062& 0.063&0.055\\
        \bottomrule
    \end{tabular}
    \caption{Attackable Sample Detection (F1) in matched setting.}
    \label{tab:m-a}
\end{table}

\begin{table}[htb!]
    \centering
    \small
        \begin{tabular}{ll|ccc}
    \toprule
       Setting  & &conf-s & conf-u & deep  \\ \midrule
        \multirow{2}{*}{all} & sst &0.448&0.449&0.476\\
        & twitter & 0.099 &0.102 &0.220\\ \midrule
        \multirow{2}{*}{uni} & sst &0.165 & 0.156 & 0.302\\
        & twitter& 0.025&0.028&0.091\\ \midrule
        \multirow{2}{*}{spec} & sst &0.340&0.340&0348\\
        & twitter& 0.088& 0.082&0.206\\ \midrule
        \multirow{2}{*}{vspec} & sst &0.126 &0.125 & 0.123\\
        & twitter& 0.025&0.015&0.053\\
        \bottomrule
    \end{tabular}
    \caption{Robust Sample Detection in matched setting.}
    \label{tab:m-r}
\end{table}

\begin{figure*}[htb!]
     \centering
     \begin{subfigure}[b]{0.24\textwidth}
         \centering
         \includegraphics[width=\textwidth]{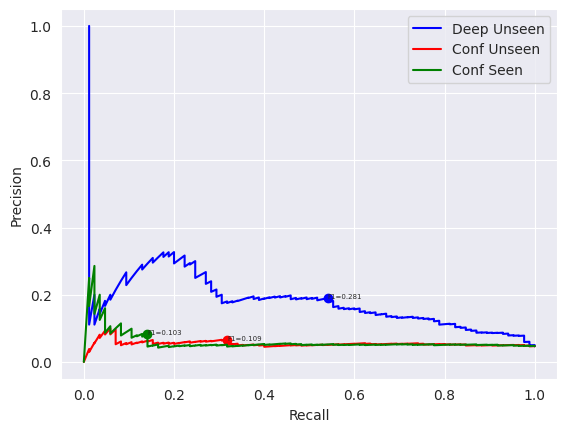}
        \caption{sst-att}
     \end{subfigure}
     \begin{subfigure}[b]{0.24\textwidth}
         \centering
         \includegraphics[width=\textwidth]{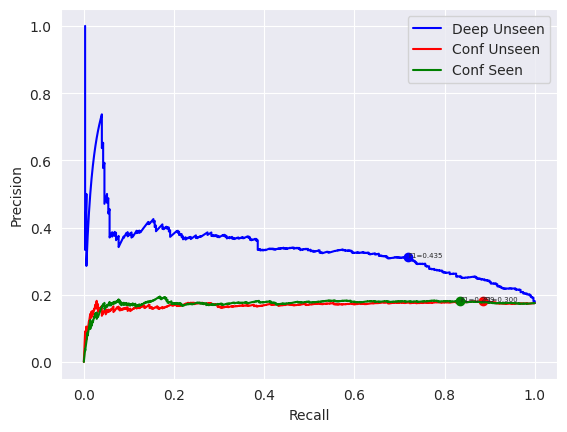}
        \caption{twitter-att}
     \end{subfigure}
 \begin{subfigure}[b]{0.24\textwidth}
         \centering
         \includegraphics[width=\textwidth]{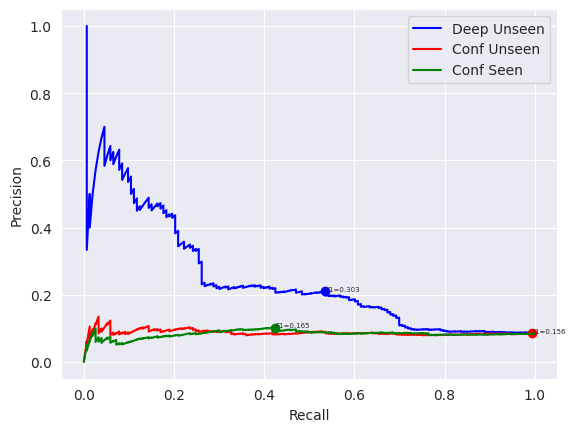}
        \caption{sst-robust}
     \end{subfigure}
     \hfill
     \begin{subfigure}[b]{0.24\textwidth}
         \centering
         \includegraphics[width=\textwidth]{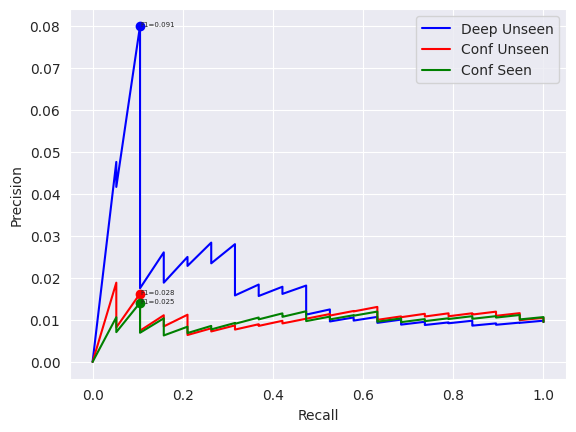}
        \caption{twitter-robust}
     \end{subfigure}
        \caption{P-R curves for detecting \textit{universal} attackable/robust samples.}
        \label{fig:pr}
\end{figure*}

Figure \ref{fig:pr}(a-b) presents the full precision-recall curves (as described in Section \ref{sec:design}) for detecting attackable samples in the \textit{uni} evaluation setting, which the deep-learning based detector has been designed for. It is evident that for a large range of operating points, the deep detection method dominates and is thus truly a useful method for identifying attackable samples. Figure \ref{fig:pr}(c-d) presents the equivalent precision-recall curves for detecting robust samples. Here, although the deep-learning method still dominates over the uncertainty-based detectors, the differences are less significant. Overall, it can be argued that this deep learning-based attackability detector is better at capturing the features of the most attackable and robust samples in a dataset than standard uncertainty based methods.

Next we want to consider the \textit{unmatched} setting, where the aim is to identify the attackable/robust samples in the test data, where the perturbation sizes for each sample are calculated using the \textit{unknown} bae attack method. Referring to Table \ref{tab:constraints}, the bae attack method has only one distance-based constraint (USE cosine distance) and so relative to the tf method with two distance based constraints, it is expected that with the definition of a sample's perturbation size, $\hat\delta^{(k)}_n$ (Equation \ref{eqn:nlp-pert}), the bae attack method will have much smaller perturbation sizes than the tf perturbation size. This is demonstrated in Figure \ref{fig:sweep-method}. Hence, for the bae attack to have a comparable number of attackable samples,  the definition of the attackable threshold is adjusted to $\epsilon_a=0.03$ and robustness threshold is kept at $\epsilon_r=0.35$. Table \ref{tab:u} gives the F1 scores for detecting universal attackable/robust samples in the unmatched uni evaluation setting. In contrast to observations made in the image domain~\citep{raina2023identifying}, here it appears that the deep detector fails to do any better than the uncertainty based detectors in identifying the attackable samples~\footnote{Interestingly, the deep detector does demonstrate some portability in identifying the most robust samples in the \textit{uni} setting, suggesting that the robust samples are similar across different attack methods.}. This suggests that the deep detector perhaps does not port over well to unknown attack methods (bae in this case) for NLP. The next section analyzes this observation further.

\begin{table}[htb!]
    \centering
    \small
    \begin{tabular}{ll|ccc}
    \toprule
        Uni setting & &conf-s & conf-u & deep  \\ \midrule
        \multirow{2}{*}{Attackable} & sst &0.555&0.555&0.555\\
        & twitter& 0.583& 0.582&0.582\\ \midrule
        \multirow{2}{*}{Robust} & sst & 0.02&0.129&0.250 \\
        & twitter& 0.001& 0.001&0.002\\
        \bottomrule
    \end{tabular}
    \caption{Sample detection (unmatched setting).}
    \label{tab:u}
\end{table}

\subsection{Portability Analysis} \label{sec:exp-port}

In the above results it is shown that a deep-learning based method performs significantly better than uncertainty-based methods in identifying attackable/robust samples for an unseen target model with a known attack method (tf), but when used to identify samples for an unknown attack method (bae), it fails to port across (for attackable sample detection). This section aims to understand this observation in greater detail. First, for each model and dataset, the known tf attack and the unknown bae attacks were used to rank samples in the validation set by the minimum perturbation size, $\hat\delta_n$. In all cases the Spearman Rank correlation is lower than 0.2 for sst and twitter (Table \ref{tab:corr}). Hence it is not surprising that the results from the matched setting do not port easily to the unmatched setting.

\begin{table}[htb!]
    \centering
    \small
    \begin{tabular}{lrrr}
    \toprule
         & bert & roberta & xlnet \\\midrule
        sst & 0.059 & 0.123&0.165\\
        twitter & 0.069 & 0.026 & 0.087\\
        \bottomrule
    \end{tabular}
    \caption{Spearman rank correlation (tf, bae).}
    \label{tab:corr}
\end{table}

To attempt to understand the lack of agreement in sample perturbation sizes between the bae and tf attack methods, we consider two further nlp attack methods: iga and pwws. For each attack method, we use the default imperceptibility constraints (pre-transformation and distance-based constraints indicated in Table \ref{tab:constraints}) and assess how effective these methods are in attacking the sst test set for each model. The results are presented in Table \ref{tab:fool}, where fooling rate is the fraction of correctly classified samples that are mis-classified after the adversarial attack. The final row considers the union of the different attack methods, where a successful attack by any one of the attack methods counts as a successful attack. It is surprising to note that although an individual attack method can achieve a fooling rate around 80\%, the union of attack methods is nearer 100\%. This demonstrates that different attack methods are able to attack a different set of samples, further highlighting that attackability/robustness of a sample is heavily dependent on the attack method.

\begin{table}[htb!]
    \centering
    \small
    \begin{tabular}{lccc|c}
    \toprule
         & \multicolumn{4}{c}{Fooling Rate (\%) } \\
         Attack & bert  & xlnet & roberta & electra\\ \midrule
       tf, $t$ & 80.7  & 79.1 & 85.4 & 76.1\\
       bae, $b$ & 63.9  & 60.8 & 65.3 & 60.7\\
       pwws, $p$ & 78.2  & 70.8 & 74.9 & 73.3\\
       iga, $i$ & 80.6 & 74.4 & 77.0 & 73.9\\\midrule
       $t\cup b\cup p\cup i$ & 96.1  & 98.1 & 98.0 & 97.3\\
       \bottomrule
    \end{tabular}
    \caption{Fooling rates with default constraints for attack methods}
    \label{tab:fool}
\end{table}

The interplay of sample attackability and the selected attack method can perhaps be explained by considering the imperceptibility constraints for each attack method. Equation \ref{eqn:pre} proposes the notion of an available set, $\mathcal A$ of possible adversarial examples that can exist for a specific source sample, $\mathbf x$, given the pre-transformation imperceptibility constraints. From Table \ref{tab:constraints} it is clear that the different attack methods have a different set of pre-transformation constraints, which suggests that each attack method can have non-overlapping available sets for a particular sample, $\mathbf{x}$, e.g. $\mathcal A^{\text{tf}}\neq \mathcal A^{\text{bae}}$. Hence, the smallest perturbation (as per the distance-based constraint) for a particular sample (Equation \ref{eqn:nlp-pert}) can change significantly across attack methods, as there is simply a different set of available adversarial examples. Hence, it can be argued that an inconsistency in sample attackability across nlp adversarial attack methods is a consequence of the differences in the pre-transformation imperceptibility constraints.

\section{Conclusions}

Little research has sought to determine the level of vulnerability of individual samples to an adversarial attack in natural language processing (NLP) tasks. This work formally extends the definitions of sample attackability to the field of NLP. It is demonstrated that uncertainty-based approaches are insufficient in characterising the most attackable and the most robust samples in a dataset. Instead, a deep-learning based detector can be used to effectively to identify these attackable/robust samples for an unseen dataset and more powerfully for an unseen target model. However, it is also observed that different attack methods in natural language have a different set of imperceptibility constraints, leading to a lack of consistency in determining sample attackability across different attack methods. As a consequence, the success of a deep-learning based attackability detector is limited to the attack method it is trained with.

\section{Limitations}

This work introduced a powerful attackability detector but also demonstrated that its success is limited to a \textit{matched} setting, where the same attack method is used in both training and evaluation of the detector. A second limitation with this work is that all experiments were carried out on natural language classification tasks. It would be useful in the future to extend these experiments to sequence-to-sequence tasks to have a more comprehensive set of results.

\section{Ethics and Broader Impact}

Adversarial attacks by nature can be of ethical concern, as malicious users can exploit theoretical adversarial attack literature to develop harmful tools to mis-use deployed deep learning systems. However this work does not aim to propose any new adversarial attack techniques, but instead considers a method to identify the most vulnerable/attackable samples. Hence, there is no perceived ethical concern related to this specific piece of work.

\section{Acknowledgements}

This paper reports on research supported by Cambridge University
Press \& Assessment (CUP\&A), a department of The Chancellor, Masters, and Scholars of the University of Cambridge.

\bibliography{custom}

\begin{thebibliography}{36}
\expandafter\ifx\csname natexlab\endcsname\relax\def\natexlab#1{#1}\fi

\bibitem[{Abdullah and Ahmet(2022)}]{10.1145/3548772}
Tariq Abdullah and Ahmed Ahmet. 2022.
\newblock \href {https://doi.org/10.1145/3548772} {Deep learning in sentiment
  analysis: Recent architectures}.
\newblock \emph{ACM Comput. Surv.}, 55(8).

\bibitem[{Alzantot et~al.(2018)Alzantot, Sharma, Elgohary, Ho, Srivastava, and
  Chang}]{DBLP:journals/corr/abs-1804-07998}
Moustafa Alzantot, Yash Sharma, Ahmed Elgohary, Bo{-}Jhang Ho, Mani~B.
  Srivastava, and Kai{-}Wei Chang. 2018.
\newblock \href {http://arxiv.org/abs/1804.07998} {Generating natural language
  adversarial examples}.
\newblock \emph{CoRR}, abs/1804.07998.

\bibitem[{Boorugu and Ramesh(2020)}]{9183355}
Ravali Boorugu and G.~Ramesh. 2020.
\newblock \href {https://doi.org/10.1109/ICIRCA48905.2020.9183355} {A survey on
  nlp based text summarization for summarizing product reviews}.
\newblock In \emph{2020 Second International Conference on Inventive Research
  in Computing Applications (ICIRCA)}, pages 352--356.

\bibitem[{Clark et~al.(2020)Clark, Luong, Le, and
  Manning}]{DBLP:journals/corr/abs-2003-10555}
Kevin Clark, Minh{-}Thang Luong, Quoc~V. Le, and Christopher~D. Manning. 2020.
\newblock \href {http://arxiv.org/abs/2003.10555} {{ELECTRA:} pre-training text
  encoders as discriminators rather than generators}.
\newblock \emph{CoRR}, abs/2003.10555.

\bibitem[{Devlin et~al.(2018)Devlin, Chang, Lee, and
  Toutanova}]{DBLP:journals/corr/abs-1810-04805}
Jacob Devlin, Ming{-}Wei Chang, Kenton Lee, and Kristina Toutanova. 2018.
\newblock \href {http://arxiv.org/abs/1810.04805} {{BERT:} pre-training of deep
  bidirectional transformers for language understanding}.
\newblock \emph{CoRR}, abs/1810.04805.

\bibitem[{Ducoffe and Precioso(2018)}]{DBLP:journals/corr/abs-1802-09841}
Melanie Ducoffe and Fr{\'{e}}d{\'{e}}ric Precioso. 2018.
\newblock \href {http://arxiv.org/abs/1802.09841} {Adversarial active learning
  for deep networks: a margin based approach}.
\newblock \emph{CoRR}, abs/1802.09841.

\bibitem[{Ebrahimi et~al.(2017)Ebrahimi, Rao, Lowd, and
  Dou}]{DBLP:journals/corr/abs-1712-06751}
Javid Ebrahimi, Anyi Rao, Daniel Lowd, and Dejing Dou. 2017.
\newblock \href {http://arxiv.org/abs/1712.06751} {Hotflip: White-box
  adversarial examples for {NLP}}.
\newblock \emph{CoRR}, abs/1712.06751.

\bibitem[{Gao et~al.(2018)Gao, Lanchantin, Soffa, and
  Qi}]{DBLP:journals/corr/abs-1801-04354}
Ji~Gao, Jack Lanchantin, Mary~Lou Soffa, and Yanjun Qi. 2018.
\newblock \href {http://arxiv.org/abs/1801.04354} {Black-box generation of
  adversarial text sequences to evade deep learning classifiers}.
\newblock \emph{CoRR}, abs/1801.04354.

\bibitem[{Garg and Ramakrishnan(2020)}]{garg-ramakrishnan-2020-bae}
Siddhant Garg and Goutham Ramakrishnan. 2020.
\newblock \href {https://doi.org/10.18653/v1/2020.emnlp-main.498} {{BAE}:
  {BERT}-based adversarial examples for text classification}.
\newblock In \emph{Proceedings of the 2020 Conference on Empirical Methods in
  Natural Language Processing (EMNLP)}, pages 6174--6181, Online. Association
  for Computational Linguistics.

\bibitem[{Goodfellow et~al.(2014)Goodfellow, Shlens, and
  Szegedy}]{https://doi.org/10.48550/arxiv.1412.6572}
Ian~J. Goodfellow, Jonathon Shlens, and Christian Szegedy. 2014.
\newblock \href {https://doi.org/10.48550/ARXIV.1412.6572} {Explaining and
  harnessing adversarial examples}.

\bibitem[{Goyal et~al.(2023)Goyal, Doddapaneni, Khapra, and
  Ravindran}]{goyal2023survey}
Shreya Goyal, Sumanth Doddapaneni, Mitesh~M. Khapra, and Balaraman Ravindran.
  2023.
\newblock \href {http://arxiv.org/abs/2203.06414} {A survey of adversarial
  defences and robustness in nlp}.

\bibitem[{Harder et~al.(2021)Harder, Pfreundt, Keuper, and
  Keuper}]{DBLP:journals/corr/abs-2103-03000}
Paula Harder, Franz{-}Josef Pfreundt, Margret Keuper, and Janis Keuper. 2021.
\newblock \href {http://arxiv.org/abs/2103.03000} {Spectraldefense: Detecting
  adversarial attacks on cnns in the fourier domain}.
\newblock \emph{CoRR}, abs/2103.03000.

\bibitem[{Herel et~al.(2022)Herel, Cisneros, and Mikolov}]{herel2022preserving}
David Herel, Hugo Cisneros, and Tomas Mikolov. 2022.
\newblock \href {http://arxiv.org/abs/2211.04205} {Preserving semantics in
  textual adversarial attacks}.

\bibitem[{Jin et~al.(2019)Jin, Jin, Zhou, and
  Szolovits}]{DBLP:journals/corr/abs-1907-11932}
Di~Jin, Zhijing Jin, Joey~Tianyi Zhou, and Peter Szolovits. 2019.
\newblock \href {http://arxiv.org/abs/1907.11932} {Is {BERT} really robust?
  natural language attack on text classification and entailment}.
\newblock \emph{CoRR}, abs/1907.11932.

\bibitem[{Kim et~al.(2021)Kim, Tack, Shin, and Hwang}]{Kim2021EntropyWA}
Minseong Kim, Jihoon Tack, Jinwoo Shin, and Sung~Ju Hwang. 2021.
\newblock Entropy weighted adversarial training.
\newblock In \emph{ICML 2021 Workshop Adversarial Machine Learning}.

\bibitem[{Li et~al.(2020{\natexlab{a}})Li, Zhang, Peng, Chen, Brockett, Sun,
  and Dolan}]{DBLP:journals/corr/abs-2009-07502}
Dianqi Li, Yizhe Zhang, Hao Peng, Liqun Chen, Chris Brockett, Ming{-}Ting Sun,
  and Bill Dolan. 2020{\natexlab{a}}.
\newblock \href {http://arxiv.org/abs/2009.07502} {Contextualized perturbation
  for textual adversarial attack}.
\newblock \emph{CoRR}, abs/2009.07502.

\bibitem[{Li et~al.(2018)Li, Ji, Du, Li, and
  Wang}]{DBLP:journals/corr/abs-1812-05271}
Jinfeng Li, Shouling Ji, Tianyu Du, Bo~Li, and Ting Wang. 2018.
\newblock \href {http://arxiv.org/abs/1812.05271} {Textbugger: Generating
  adversarial text against real-world applications}.
\newblock \emph{CoRR}, abs/1812.05271.

\bibitem[{Li et~al.(2020{\natexlab{b}})Li, Ma, Guo, Xue, and
  Qiu}]{DBLP:journals/corr/abs-2004-09984}
Linyang Li, Ruotian Ma, Qipeng Guo, Xiangyang Xue, and Xipeng Qiu.
  2020{\natexlab{b}}.
\newblock \href {http://arxiv.org/abs/2004.09984} {{BERT-ATTACK:} adversarial
  attack against {BERT} using {BERT}}.
\newblock \emph{CoRR}, abs/2004.09984.

\bibitem[{Liu et~al.(2019)Liu, Ott, Goyal, Du, Joshi, Chen, Levy, Lewis,
  Zettlemoyer, and Stoyanov}]{DBLP:journals/corr/abs-1907-11692}
Yinhan Liu, Myle Ott, Naman Goyal, Jingfei Du, Mandar Joshi, Danqi Chen, Omer
  Levy, Mike Lewis, Luke Zettlemoyer, and Veselin Stoyanov. 2019.
\newblock \href {http://arxiv.org/abs/1907.11692} {Roberta: {A} robustly
  optimized {BERT} pretraining approach}.
\newblock \emph{CoRR}, abs/1907.11692.

\bibitem[{Morris et~al.(2020)Morris, Lifland, Lanchantin, Ji, and
  Qi}]{morris-etal-2020-reevaluating}
John Morris, Eli Lifland, Jack Lanchantin, Yangfeng Ji, and Yanjun Qi. 2020.
\newblock \href {https://doi.org/10.18653/v1/2020.findings-emnlp.341}
  {Reevaluating adversarial examples in natural language}.
\newblock In \emph{Findings of the Association for Computational Linguistics:
  EMNLP 2020}, pages 3829--3839, Online. Association for Computational
  Linguistics.

\bibitem[{Pruthi et~al.(2019)Pruthi, Dhingra, and
  Lipton}]{DBLP:journals/corr/abs-1905-11268}
Danish Pruthi, Bhuwan Dhingra, and Zachary~C. Lipton. 2019.
\newblock \href {http://arxiv.org/abs/1905.11268} {Combating adversarial
  misspellings with robust word recognition}.
\newblock \emph{CoRR}, abs/1905.11268.

\bibitem[{Qian et~al.(2022)Qian, Huang, Wang, and Zhang}]{qian2022survey}
Zhuang Qian, Kaizhu Huang, Qiu-Feng Wang, and Xu-Yao Zhang. 2022.
\newblock \href {http://arxiv.org/abs/2203.14046} {A survey of robust
  adversarial training in pattern recognition: Fundamental, theory, and
  methodologies}.

\bibitem[{Raina and Gales(2022)}]{Raina_2022}
Vyas Raina and Mark Gales. 2022.
\newblock \href {https://doi.org/10.18653/v1/2022.naacl-main.281}
  {Residue-based natural language adversarial attack detection}.
\newblock In \emph{Proceedings of the 2022 Conference of the North American
  Chapter of the Association for Computational Linguistics: Human Language
  Technologies}. Association for Computational Linguistics.

\bibitem[{Raina and Gales(2023)}]{raina2023identifying}
Vyas Raina and Mark Gales. 2023.
\newblock \href {http://arxiv.org/abs/2301.12896} {Identifying adversarially
  attackable and robust samples}.

\bibitem[{Ren et~al.(2020)Ren, Xiao, Chang, Huang, Li, Chen, and
  Wang}]{DBLP:journals/corr/abs-2009-00236}
Pengzhen Ren, Yun Xiao, Xiaojun Chang, Po{-}Yao Huang, Zhihui Li, Xiaojiang
  Chen, and Xin Wang. 2020.
\newblock \href {http://arxiv.org/abs/2009.00236} {A survey of deep active
  learning}.
\newblock \emph{CoRR}, abs/2009.00236.

\bibitem[{Ren et~al.(2019)Ren, Deng, He, and Che}]{ren-etal-2019-generating}
Shuhuai Ren, Yihe Deng, Kun He, and Wanxiang Che. 2019.
\newblock \href {https://doi.org/10.18653/v1/P19-1103} {Generating natural
  language adversarial examples through probability weighted word saliency}.
\newblock In \emph{Proceedings of the 57th Annual Meeting of the Association
  for Computational Linguistics}, pages 1085--1097, Florence, Italy.
  Association for Computational Linguistics.

\bibitem[{Saravia et~al.(2018)Saravia, Liu, Huang, Wu, and
  Chen}]{saravia-etal-2018-carer}
Elvis Saravia, Hsien-Chi~Toby Liu, Yen-Hao Huang, Junlin Wu, and Yi-Shin Chen.
  2018.
\newblock \href {https://doi.org/10.18653/v1/D18-1404} {{CARER}: Contextualized
  affect representations for emotion recognition}.
\newblock In \emph{Proceedings of the 2018 Conference on Empirical Methods in
  Natural Language Processing}, pages 3687--3697, Brussels, Belgium.
  Association for Computational Linguistics.

\bibitem[{Socher et~al.(2013)Socher, Perelygin, Wu, Chuang, Manning, Ng, and
  Potts}]{socher-etal-2013-recursive}
Richard Socher, Alex Perelygin, Jean Wu, Jason Chuang, Christopher~D. Manning,
  Andrew Ng, and Christopher Potts. 2013.
\newblock \href {https://www.aclweb.org/anthology/D13-1170} {Recursive deep
  models for semantic compositionality over a sentiment treebank}.
\newblock In \emph{Proceedings of the 2013 Conference on Empirical Methods in
  Natural Language Processing}, pages 1631--1642, Seattle, Washington, USA.
  Association for Computational Linguistics.

\bibitem[{Sun and Wang(2010)}]{5581075}
Li-Li Sun and Xi-Zhao Wang. 2010.
\newblock \href {https://doi.org/10.1109/ICMLC.2010.5581075} {A survey on
  active learning strategy}.
\newblock In \emph{2010 International Conference on Machine Learning and
  Cybernetics}, volume~1, pages 161--166.

\bibitem[{Tan et~al.(2020)Tan, Joty, Kan, and Socher}]{tan-etal-2020-morphin}
Samson Tan, Shafiq Joty, Min-Yen Kan, and Richard Socher. 2020.
\newblock \href {https://doi.org/10.18653/v1/2020.acl-main.263} {It{'}s
  morphin{'} time! {C}ombating linguistic discrimination with inflectional
  perturbations}.
\newblock In \emph{Proceedings of the 58th Annual Meeting of the Association
  for Computational Linguistics}, pages 2920--2935, Online. Association for
  Computational Linguistics.

\bibitem[{Vaswani et~al.(2017)Vaswani, Shazeer, Parmar, Uszkoreit, Jones,
  Gomez, Kaiser, and Polosukhin}]{DBLP:journals/corr/VaswaniSPUJGKP17}
Ashish Vaswani, Noam Shazeer, Niki Parmar, Jakob Uszkoreit, Llion Jones,
  Aidan~N. Gomez, Lukasz Kaiser, and Illia Polosukhin. 2017.
\newblock \href {http://arxiv.org/abs/1706.03762} {Attention is all you need}.
\newblock \emph{CoRR}, abs/1706.03762.

\bibitem[{Wang et~al.(2019)Wang, Jin, and
  He}]{DBLP:journals/corr/abs-1909-06723}
Xiaosen Wang, Hao Jin, and Kun He. 2019.
\newblock \href {http://arxiv.org/abs/1909.06723} {Natural language adversarial
  attacks and defenses in word level}.
\newblock \emph{CoRR}, abs/1909.06723.

\bibitem[{Xu et~al.(2022)Xu, Zhang, Zheng, Li, Hsieh, Chang, and
  Huang}]{xu-etal-2022-towards}
Jianhan Xu, Cenyuan Zhang, Xiaoqing Zheng, Linyang Li, Cho-Jui Hsieh, Kai-Wei
  Chang, and Xuanjing Huang. 2022.
\newblock \href {https://doi.org/10.18653/v1/2022.findings-acl.134} {Towards
  adversarially robust text classifiers by learning to reweight clean
  examples}.
\newblock In \emph{Findings of the Association for Computational Linguistics:
  ACL 2022}, pages 1694--1707, Dublin, Ireland. Association for Computational
  Linguistics.

\bibitem[{Yang et~al.(2020)Yang, Wang, and
  Chu}]{DBLP:journals/corr/abs-2002-07526}
Shuoheng Yang, Yuxin Wang, and Xiaowen Chu. 2020.
\newblock \href {http://arxiv.org/abs/2002.07526} {A survey of deep learning
  techniques for neural machine translation}.
\newblock \emph{CoRR}, abs/2002.07526.

\bibitem[{Yang et~al.(2019)Yang, Dai, Yang, Carbonell, Salakhutdinov, and
  Le}]{DBLP:journals/corr/abs-1906-08237}
Zhilin Yang, Zihang Dai, Yiming Yang, Jaime~G. Carbonell, Ruslan Salakhutdinov,
  and Quoc~V. Le. 2019.
\newblock \href {http://arxiv.org/abs/1906.08237} {Xlnet: Generalized
  autoregressive pretraining for language understanding}.
\newblock \emph{CoRR}, abs/1906.08237.

\bibitem[{Zeng et~al.(2020)Zeng, Zhu, Goldstein, and
  Huang}]{DBLP:journals/corr/abs-2010-12989}
Huimin Zeng, Chen Zhu, Tom Goldstein, and Furong Huang. 2020.
\newblock \href {http://arxiv.org/abs/2010.12989} {Are adversarial examples
  created equal? {A} learnable weighted minimax risk for robustness under
  non-uniform attacks}.
\newblock \emph{CoRR}, abs/2010.12989.

\end{thebibliography}
\bibliographystyle{acl_natbib}

\appendix
\newpage

\section*{Appendix}

\section{Full set of empirical results}
\label{sec:appendix}

\begin{figure}[htb!]
     \centering
     \begin{subfigure}[b]{0.45\textwidth}
         \centering
         \includegraphics[width=\textwidth]{Figures/sweep_textfooler.png}
        \caption{textfooler}
         \label{fig:temp1}
     \end{subfigure}
     \hfill
     \begin{subfigure}[b]{0.45\textwidth}
         \centering
         \includegraphics[width=\textwidth]{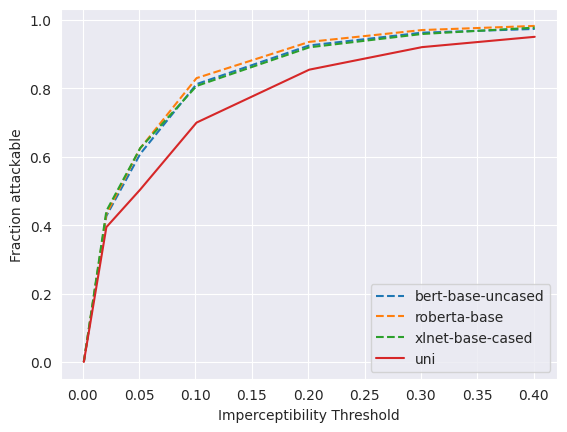}
        \caption{bae}
         \label{fig:temp2}
     \end{subfigure}
        \caption{Fraction of samples classed as adversarially attackable across model architecture with increasing imperceptibility threshold as per distance-based constraint (sst).}
        \label{fig:sweep-model}
\end{figure}

\begin{figure}[htb!]
     \centering
     \begin{subfigure}[b]{0.45\textwidth}
         \centering
         \includegraphics[width=\textwidth]{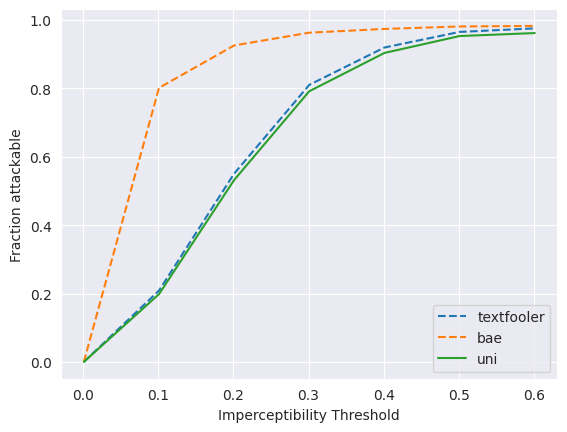}
        \caption{bert}
         \label{fig:temp5}
     \end{subfigure}
     \hfill
     \begin{subfigure}[b]{0.45\textwidth}
         \centering
         \includegraphics[width=\textwidth]{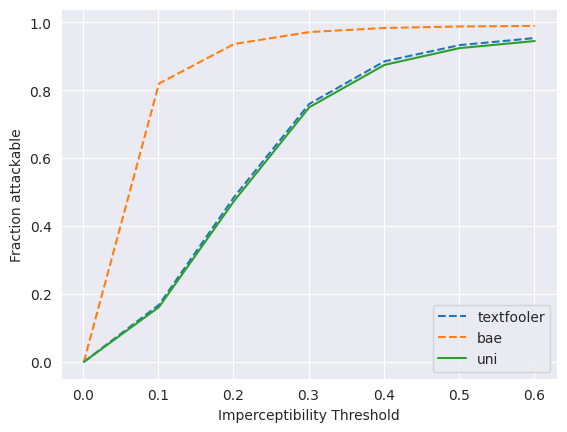}
        \caption{roberta}
         \label{fig:temp6}
     \end{subfigure}
     \hfill
     \begin{subfigure}[b]{0.45\textwidth}
         \centering
         \includegraphics[width=\textwidth]{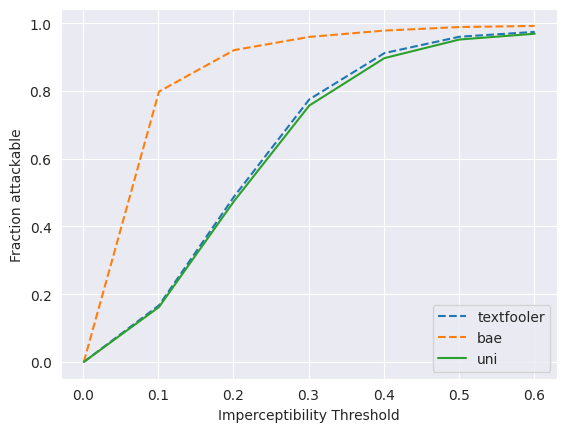}
        \caption{xlnet}
         \label{fig:temp7}
     \end{subfigure}
        \caption{Fraction of samples classed as adversarially attackable across attack method with increasing imperceptibility threshold as per distance-based constraint (sst).}
        \label{fig:sweep-method}
\end{figure}

\begin{figure}[htb!]
     \centering
     \begin{subfigure}[b]{0.45\textwidth}
         \centering
         \includegraphics[width=\textwidth]{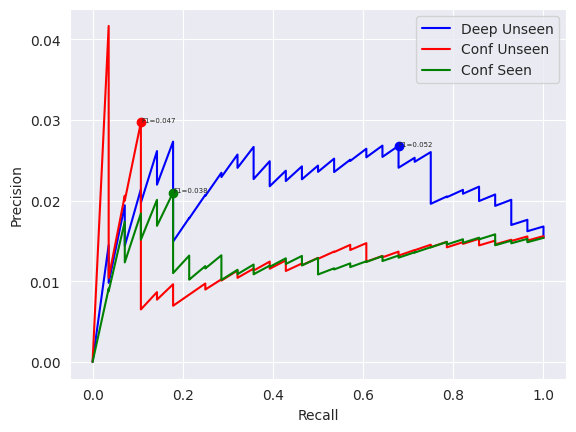}
        \caption{Very Specific}
         \label{fig:temp8}
     \end{subfigure}
     \hfill
     \begin{subfigure}[b]{0.45\textwidth}
         \centering
         \includegraphics[width=\textwidth]{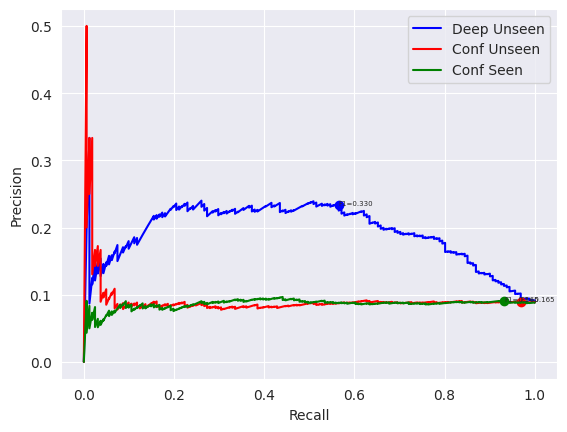}
        \caption{Specific}
         \label{fig:temp9}
     \end{subfigure}
     \hfill
     \begin{subfigure}[b]{0.45\textwidth}
         \centering
         \includegraphics[width=\textwidth]{Figures/sst_textfooler_uni.png}
        \caption{Universal}
         \label{fig:temp10}
     \end{subfigure}
     \hfill
     \begin{subfigure}[b]{0.45\textwidth}
         \centering
         \includegraphics[width=\textwidth]{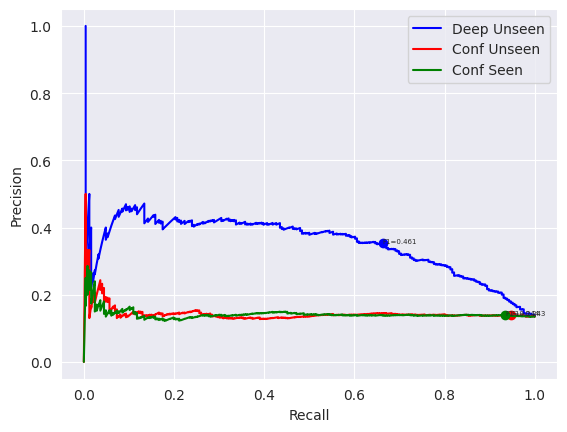}
        \caption{All}
         \label{fig:temp11}
     \end{subfigure}
        \caption{PR curves: Attackable Sample Detection (sst)}
        \label{fig:pr-att}
\end{figure}

\begin{figure}[htb!]
     \centering
     \begin{subfigure}[b]{0.45\textwidth}
         \centering
         \includegraphics[width=\textwidth]{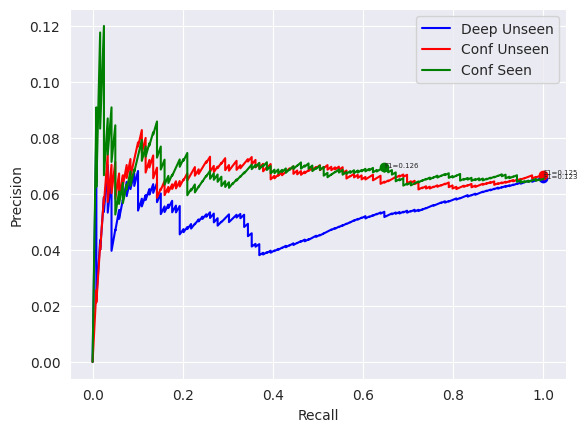}
        \caption{Very Specific}
         \label{fig:temp12}
     \end{subfigure}
     \hfill
     \begin{subfigure}[b]{0.45\textwidth}
         \centering
         \includegraphics[width=\textwidth]{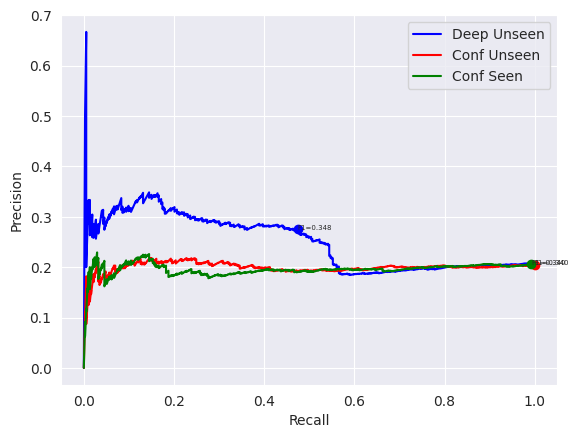}
        \caption{Specific}
         \label{fig:temp13}
     \end{subfigure}
     \hfill
     \begin{subfigure}[b]{0.45\textwidth}
         \centering
         \includegraphics[width=\textwidth]{Figures/robust_sst_textfooler_uni.png}
        \caption{Universal}
         \label{fig:temp14}
     \end{subfigure}
     \hfill
     \begin{subfigure}[b]{0.45\textwidth}
         \centering
         \includegraphics[width=\textwidth]{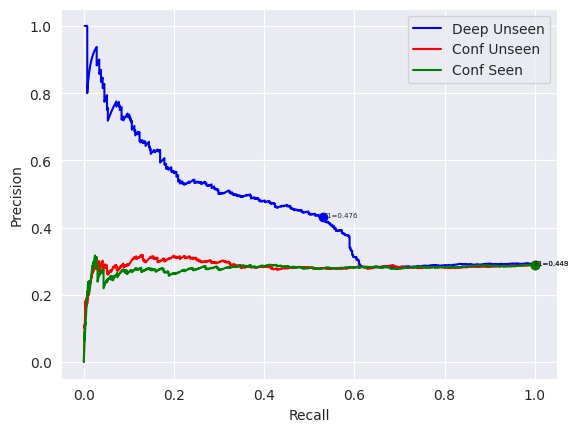}
        \caption{All}
         \label{fig:pr-robust}
     \end{subfigure}
        \caption{PR curves: Robust Sample Detection (sst)}
        \label{fig:detect-final}
\end{figure}

\end{document}